\documentclass[journal]{IEEEtran}
\usepackage{cite}
\usepackage{mathrsfs}
\usepackage{amsmath}
\usepackage{graphicx}
\usepackage{subfigure}
\usepackage{amsfonts}
\usepackage{bm}
\usepackage{arydshln}
\usepackage{algorithm}  
\usepackage{algorithmicx}  
\usepackage{algpseudocode}  
\usepackage{booktabs}
\usepackage[marginal]{footmisc}
\usepackage{orcidlink}
\usepackage{subcaption}

\floatname{algorithm}{Algorithm}  

\usepackage{lineno,hyperref}

\ifCLASSINFOpdf
\else
\fi
\begin{document}
%
\title{UPETrack: Unidirectional Position Estimation for Tracking Occluded Deformable Linear Objects}
%
%
%

\author{  
        Fan~Wu$^{\orcidlink{0009-0007-9496-6373}}$,
        Chenguang~Yang$^{\orcidlink{0000-0001-5255-5559}}$,
        Haibin~Yang$^{\orcidlink{0000-0003-2871-610X}}$,
        Shuo~Wang, 
        Yanrui~Xu, 
        Xing~Zhou$^{\orcidlink{0000-0003-2871-610X}}$,
        Meng~Gao$^{\orcidlink{0000-0003-2871-610X}}$,
        Yaoqi~Xian$^{\orcidlink{0000-0002-4404-4473}}$,
        Zhihong~Zhu$^{\orcidlink{0000-0003-2871-610X}}$,
        Shifeng~Huang$^{\orcidlink{0000-0003-2871-610X}}$,~\IEEEmembership{Member, IEEE}

\thanks{\indent Fan Wu and Zhihong Zhu are with the Huazhong University of Science and Technology, Wuhan 430074, China.}
\thanks{\indent Haibin Yang, Xing Zhou, Meng Gao, and Yaoqi Xian are with the Foshan Institute of Intelligent Equipment Technology, Foshan 528234, China (e-mail: qunw@hzncc.com; zhouxing@hzncc.com; rwxyrwx@163.com; xianyaoqi@gmail.com).}
\thanks{\indent Shifeng Huang is with the School of Artificial Intelligence and Robotics, Hunan University, Changsha 410082, China (e-mail: fenghkust@gmail.com).}
}

%
%

\markboth{ }%
{Shell \MakeLowercase{\textit{et al.}}: Bare Demo of IEEEtran.cls for IEEE Journals}
%



\maketitle

\begin{abstract}

Real-time state tracking of Deformable Linear Objects (DLOs) is critical for enabling robotic 
manipulation of DLOs in industrial assembly, medical procedures, and daily-life applications. 
However, the high-dimensional configuration space, nonlinear dynamics, and frequent partial 
occlusions present fundamental barriers to robust real-time DLO tracking. To address these 
limitations, this study introduces UPETrack, a geometry-driven framework based on Unidirectional Position Estimation (UPE), 
which facilitates tracking without the requirement for physical modeling, virtual simulation, or visual markers. 
The framework operates in two phases: (1) visible segment tracking is based on a Gaussian Mixture 
Model (GMM) fitted via the Expectation Maximization (EM) algorithm, and (2) occlusion region prediction employing 
UPE algorithm we proposed. UPE leverages the geometric 
continuity inherent in DLO shapes and their temporal evolution patterns to derive a closed-form 
positional estimator through three principal mechanisms: (i) local linear combination displacement 
term, (ii) proximal linear constraint term, and (iii) historical curvature term. This analytical 
formulation allows efficient and stable estimation of occluded nodes through explicit linear combinations 
of geometric components, eliminating the need for additional iterative optimization. 
Experimental results demonstrate that UPETrack surpasses two state-of-the-art 
tracking algorithms, including TrackDLO and CDCPD2, in both positioning accuracy and computational 
efficiency.

\end{abstract}

%
\IEEEpeerreviewmaketitle

\section{Introduction}
\IEEEPARstart{W}{ith} the continuous expansion of robotic applications across diverse domains, the perception and manipulation 
technologies for deformable objects have rapidly emerged as a cutting-edge research frontier in robotics\cite{ref12}, \cite{ref28}. 
As one of the core challenges in this field, real-time visual recognition and tracking of marker-free 
Deformable Linear Objects (DLOs) demonstrate significant application value in industrial manufacturing, 
medical surgery, and service robots \cite{ref6}, \cite{ref8}, \cite{refWang2023}. Typical application scenarios encompass robotic 
wire harness assembly in industrial settings \cite{10341726}, \cite{ref2}, \cite{ref10}, precise control of suturing threads in minimally 
invasive surgery \cite{ref3}, \cite{ref7}, and autonomous cable management in domestic environments \cite{ref4}, \cite{ref5}. 
These application scenarios universally demand robotic systems to possess real-time perception 
capabilities for the dynamic states of DLOs to achieve precise manipulation 
toward target configurations. However, partial occlusions caused by environmental disturbances or 
robotic operations may result in unobservable morphological features of DLO segments, consequently 
eliciting decision-making errors or even task failures, as shown in Fig. 1(a). To address this, it is 
imperative to develop occlusion-robust real-time state tracking techniques for DLOs, which enable 
online reconstruction of their complete geometric profiles to provide reliable state observations 
for robotic decision-making, as shown in Fig. 1(b). Nevertheless, the tracking task poses 
significant technical challenges due to the inherent complexities of DLOs, including high-dimensional 
state spaces and nonlinear dynamic characteristics.

\begin{figure}[t]
  \centering
  \begin{subfigure}[]{
    \includegraphics[scale=0.115]{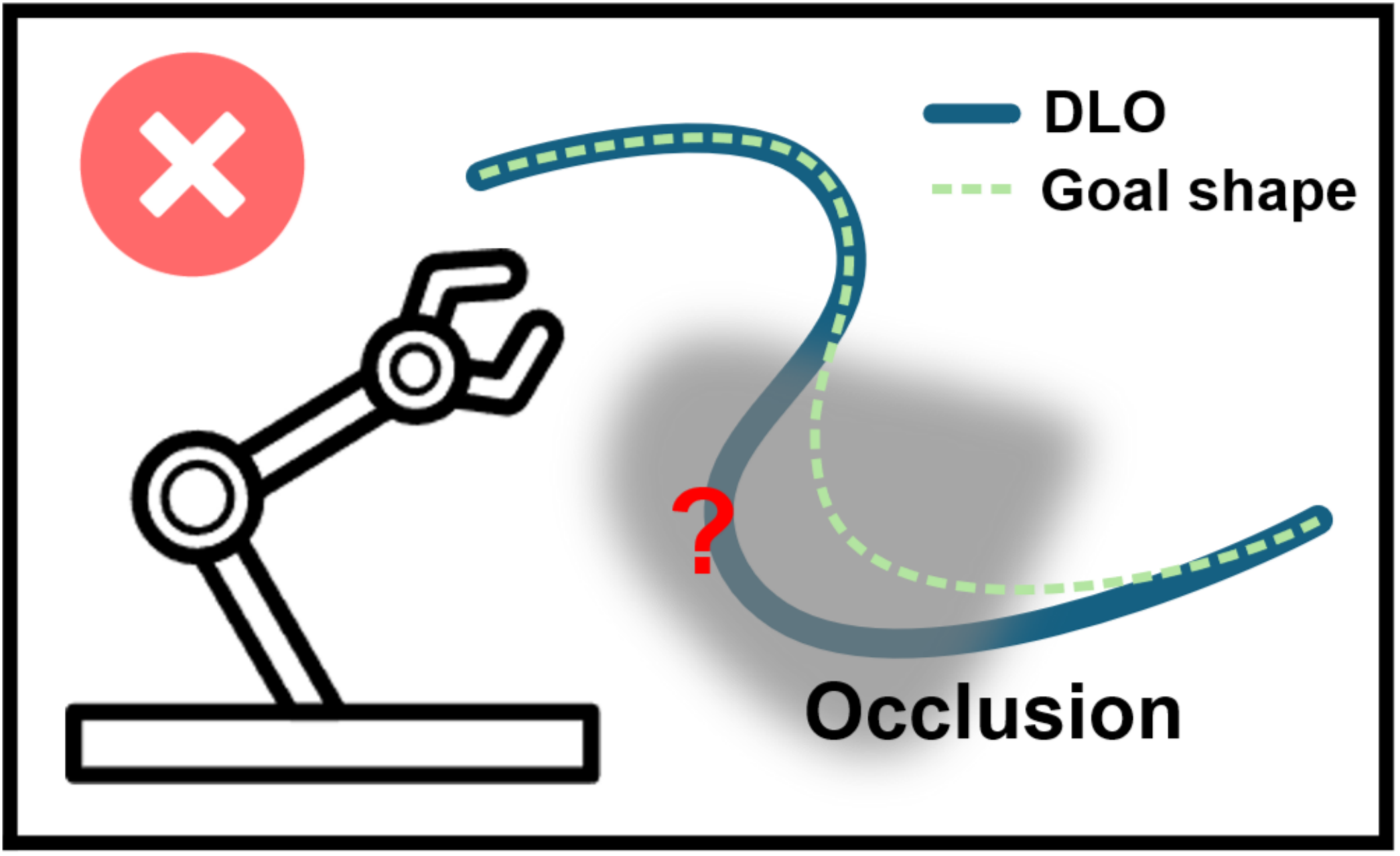}}
    \end{subfigure}
  \begin{subfigure}[]{
    \includegraphics[scale=0.115]{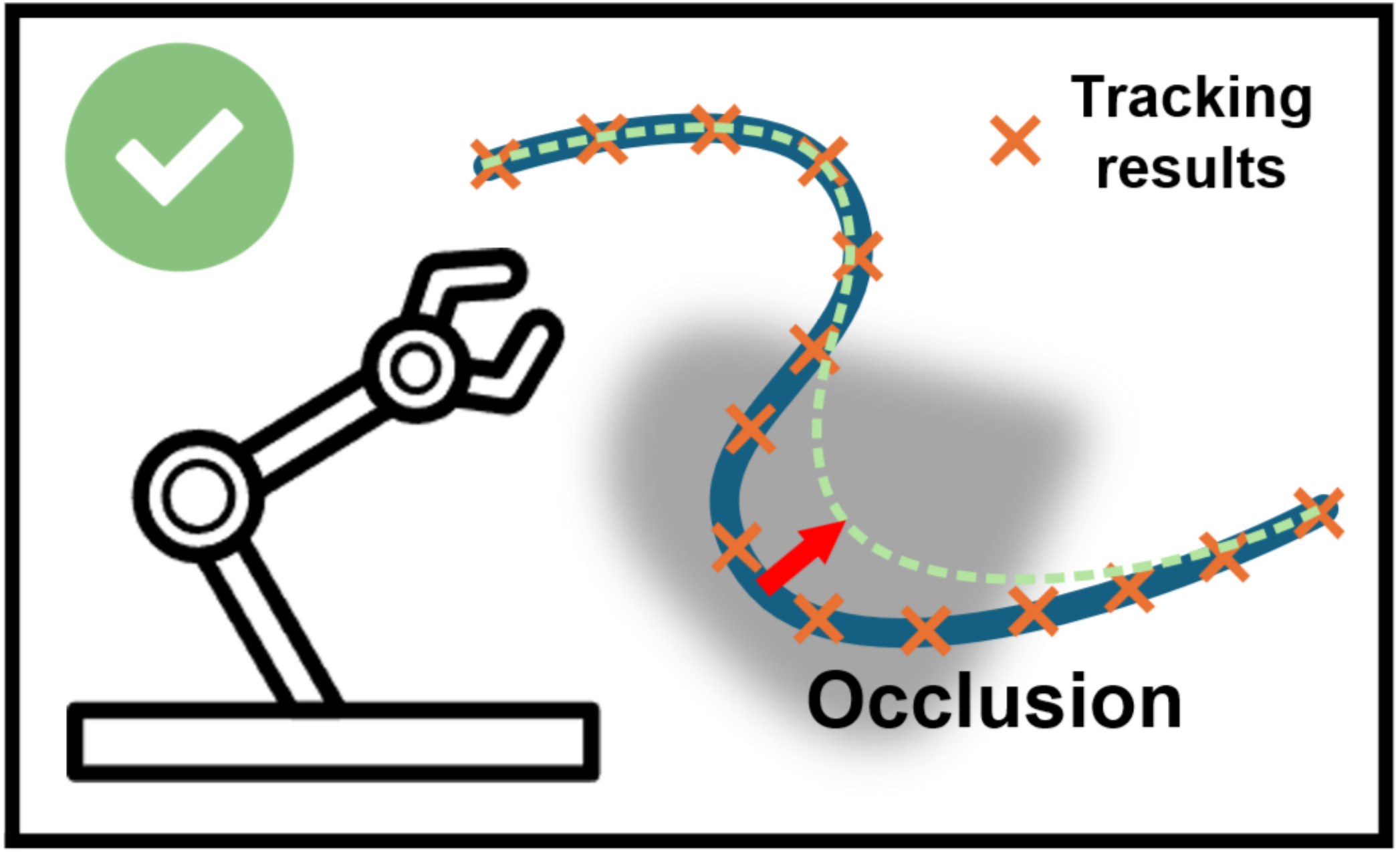}}
    \end{subfigure}
  \caption{(a) The robotic manipulation of a DLO usually involves 
  transforming the DLO from its current configuration into a goal shape. However, local occlusions 
  can compromise the completeness of the perceived shape of the DLO, severely hampering robotic 
  decision-making. (b) Therefore, it is necessary to establish tracking algorithms for real-time state estimation 
  of DLOs, enabling the reconstruction of their complete geometric configuration based on the tracking results.}
  \label{Fourier_Excitation_Advantage}
\end{figure} 

\begin{figure*}[t]
  \centering
  \includegraphics[width=0.98\textwidth]{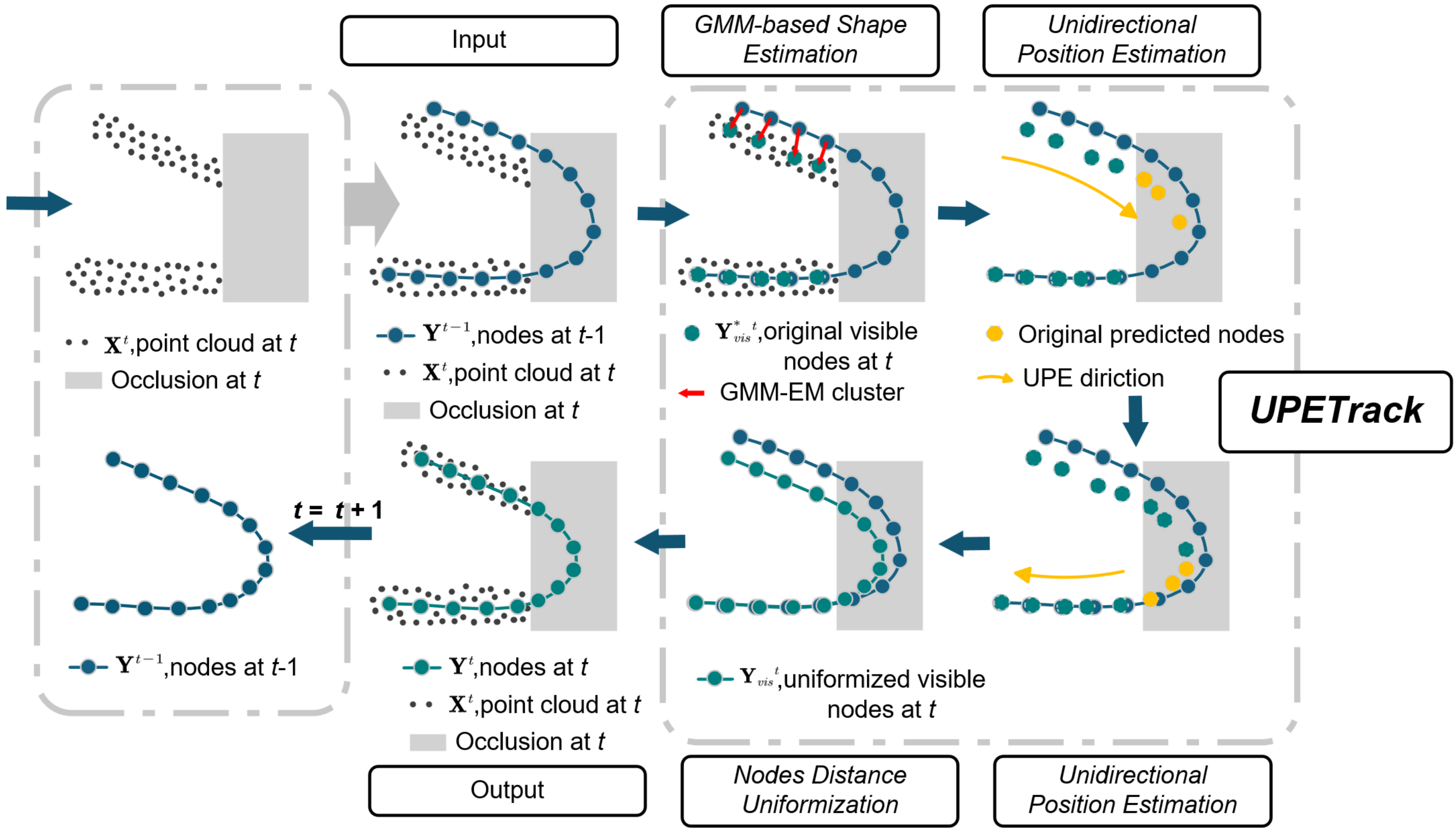} 
  \caption{UPETrack integrates the current DLO point cloud with prior temporal nodes to obtain visible segments. 
  It then employs Unidirectional Position Estimation (UPE) for occluded node localization, followed by uniform 
  resampling to generate final state estimates.}
  \label{fig:XXX}
\end{figure*}

In current research practices, DLOs are commonly modeled as $M$ nodes evenly distributed along their length, forming 
a state vector of dimension $D \times M$, where $D$ corresponds to the coordinate dimensions of each node. Previous 
methods have successfully achieved real-time detection without relying on visual markers or physical 
simulation \cite{ref21-8}, \cite{ref23-10}, \cite{ref24-11}. These approaches primarily address occlusion-related state estimation by integrating 
constraints such as Coherent Point Drift (CPD) \cite{ref14-1} and Global-Local Topology Preservation (GLTP) \cite{ref11} into 
the optimization framework of the Expectation-Maximization (EM) algorithm \cite{ref13}. Other algorithms can also 
track DLO states through physical simulation or particle filters \cite{ref26-13}. 
The majority of existing methods typically incorporate geometric constraints within iterative optimization 
frameworks, resulting in two inherent limitations: (1) increased computational 
complexity and (2) diminished convergence stability. 

To overcome these limitations, we propose UPETrack, a geometry-driven framework. Unlike most existing methods that embed geometric constraints into iterative optimization frameworks, 
UPETrack directly computes closed-form solutions for occluded node positions by leveraging the spatiotemporal geometric characteristics of DLOs, 
thereby avoiding iterative optimization within the occlusion inference stage. This approach leverages the spatiotemporal geometric regularities of 
DLOs to analytically infer occluded node positions, providing an efficient and stable strategy for real-time tracking.
Specifically, UPE combines temporal regularity of shape deformation (e.g., local linear combination displacement term and historical curvature term) 
with spatial continuity constraints (e.g., proximal linear constraint term) to achieve real-time tracking.
Experimental results demonstrate that UPETrack exhibits notable accuracy and efficiency in addressing DLO 
tracking problems under partial occlusion. The principal contributions of this work are threefold:

\begin{itemize}
  \item UPETrack achieves accurate real-time tracking of DLOs under partial occlusion without the need for  
  physical modeling, simulation, or visual markers.

  \item The proposed UPE algorithm leverages the intrinsic spatiotemporal geometric properties 
  of DLOs, including spatial continuity and temporal deformation regularity. These properties 
  are formulated as analytical constraints to govern the state of occluded nodes.

  \item A closed-form solution is developed to directly compute the positions of occluded nodes from 
  the above spatiotemporal constraints. This analytical computation bypasses iterative optimization, 
  leading to enhanced tracking stability and real-time computational efficiency.
\end{itemize}

\section{Related Work}
The non-rigid registration method based on the Coherent Point Drift (CPD) algorithm demonstrates notable 
advantages in tracking DLOs under occlusion and has attracted 
considerable research attention in recent years. The core principle lies in establishing spatial correspondence 
between observed point clouds and topological nodes representing DLO structures. 
The CPD framework achieves non-rigid point set registration through a systematic 
approach: (1) modeling the target point set as a Gaussian mixture model (GMM), (2) enforcing 
motion coherence constraints between adjacent control points via the Motion Coherence Theory (MCT) \cite{ref15-2}, 
and (3) iteratively optimizing transformation parameters using the expectation-maximization (EM) 
algorithm.

Researchers have utilized physics-based simulation methods to address the tracking of deformable objects under 
occlusion. For instance, references \cite{ref16-3}, \cite{ref17-4} and \cite{ref18-5} combined the Gaussian Mixture Model (GMM) with regularized 
Coherent Point Drift (CPD) and leveraged dynamic simulation from a physics engine to ensure that the 
estimated results adhere to the physical constraints of the objects. However, these algorithms have limitations, 
as they require prior knowledge of certain key physical parameters of the deformable objects. To address this issue, 
Jin et al. \cite{ref20-7} introduced point cloud recovery to tackle occlusion; however, it was unable to track moving 
DLOs under partial occlusion. Chi et al. \cite{ref21-8} proposed a Constrained Deformable 
CPD (CDCPD) algorithm, which improves the output of the CPD by incorporating Locally Linear Embedding (LLE) \cite{ref22-9} 
and constrained optimization, thereby reducing the dependence on physical models. Furthermore, Y. Wang et al. proposed the CDCPD2 algorithm \cite{ref23-10},
which incorporates convex constraints to prevent self-intersections and obstacle penetration, 
thereby addressing the instability in tracking performance under occlusion that was observed 
in the original method. The TrackDLO algorithm \cite{ref24-11} proposed by J. Xiang et al. applies Motion Coherence Theory (MCT) to infer the spatial 
velocities of occluded nodes, uses topological geodesic distances to track self-occluded DLOs, and introduces a 
non-Gaussian kernel to better reflect the physical properties of DLOs. 

In addition to the aforementioned approaches, 
Zhang et al. presented \cite{ref25-12} that detects keypoints from RGB images and infers their correct order using a confidence-guided matching 
strategy. The full object shape is then reconstructed via spline interpolation.  
Yang et al. \cite{ref26-13} achieved robust tracking of DLO states through the application of a particle filter in a 
low-dimensional latent space, maintaining high accuracy even under severe occlusion conditions. Lv et al. \cite{ref27-14}
proposed a dual-branch network architecture that separately exploits global and local information, along with a 
specifically designed fusion module to effectively combine the strengths of both strategies. This method 
demonstrated the capability to generate DLO state estimations that are globally smooth while maintaining local 
precision.

These existing tracking methods predominantly incorporate the geometric properties of DLOs as 
constraints into iterative optimization frameworks or physics-based simulations. 

\section{The UPETrack Framework}

The UPETrack represents the DLO shape with a collection of ordered nodes. As shown in Fig. 2,  
the $N$ points $\textbf{X}^{t}$ represent the partial observation of the target DLO at time step $t$, 
where $\textbf{X}^{t}=[\textbf{x}_{1}^{t},\textbf{x}_{2}^{t},\dots,\textbf{x}_{N}^{t}]^T \in \mathbb{R} ^{N\times D}$. 
To obtain these points from the depth image (captured by a depth sensor), we employ a simple RGB-based color thresholding approach for DLO segmentation. 
The algorithm generates $M$ state nodes $\textbf{Y} ^{t}$ 
to represent the shape of DLO, where 
$\textbf{Y} ^{t}=[\textbf{y}_{1}^{t},\textbf{y}_{2}^{t},\dots,\textbf{y}_{M}^{t}]^T \in \mathbb{R} ^{M\times D}$, and $D$ 
denotes the dimension of the points and nodes. $\textbf{X} ^{t}$ could 
be incomplete due to occlusions. Assuming the topology of the DLO remains constant, $\textbf{y}_{i}^{t}$ and $\textbf{y}_{i}^{t'}$ 
correspond to the same point at DLO for time step $t$ and $t'$.

The UPETrack is initialized using an occlusion-robust DLO detection framework that models linear 
objects within the image as smooth curves\cite{ref29}. 
Initial DLO state nodes $\textbf{Y} ^{0}$ are generated through uniform sampling along the reconstructed curve.
Following this, UPETrack aligns the predicted visible state nodes $\textbf{Y}_{vis}^{t}$ (as formally defined in Section III-A)
to the current observation $\textbf{X}^{t}$ by fitting a GMM initialized 
with $\textbf{Y}_{vis}^{t-1}$ to $\textbf{X}^{t}$ using the EM algorithm.
To address the tracking of the occluded segments of DLOs, the framework 
incorporates a Unidirectional Position Estimation (UPE) algorithm that significantly reduces computational time
complexity compared to existing approaches.

\subsection{GMM-Based Shape Estimation} 

\begin{figure}[t]
  \centering
  \includegraphics[scale=0.23]{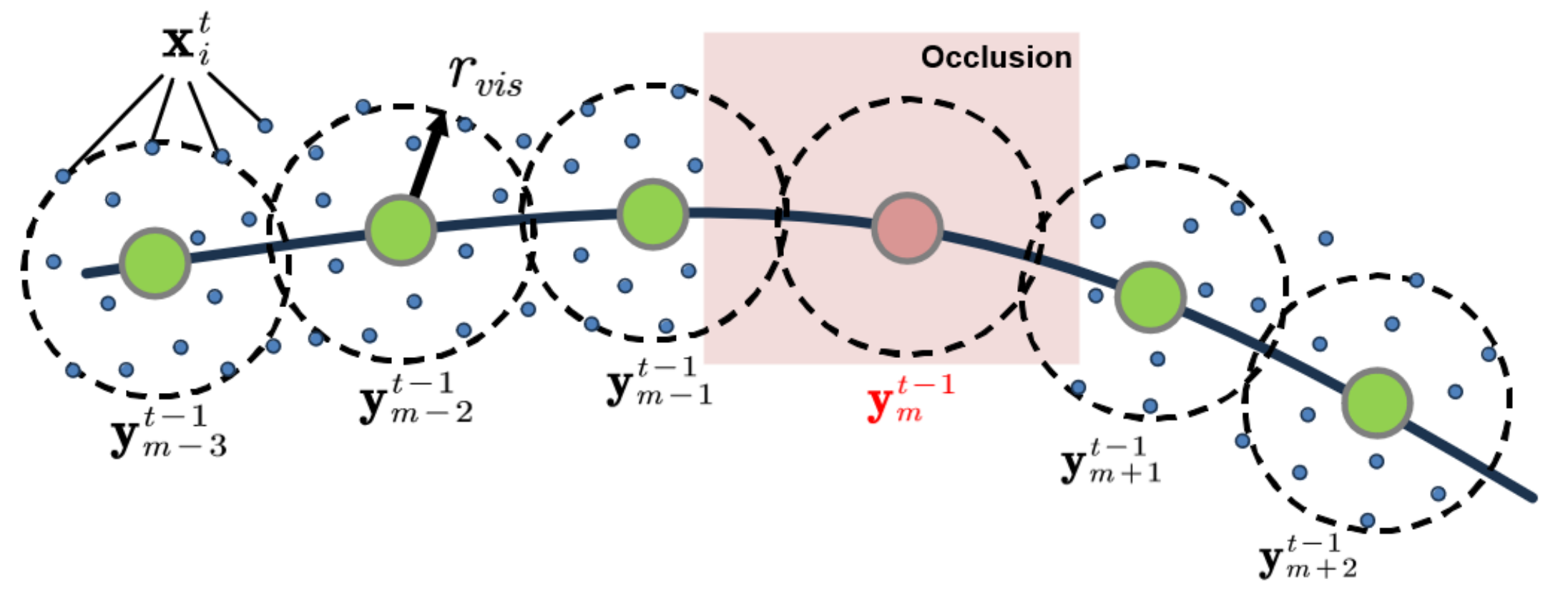}
  \caption{The algorithm determines node visibility by checking whether the number of neighboring points ($q$)
  within radius $r_{vis}$ around $\textbf{y}_{m}^{t-1}$ exceeds threshold $V_{lim}$, marking $\textbf{y}_{m}^{t}$ 
  as visible if $q\geq V_{lim}$ or occluded otherwise.}
  \label{figure}
\end{figure}

The visibility determination of state nodes constitutes a critical component of UPETrack, as illustrated in Fig. 3. 
It evaluates the visibility of a state node $\textbf{y}_{m}^{t}$ by counting the number $q$ of sampled 
points $\textbf{x}_{i}^{t}\in \textbf{X} ^{t}$ that satisfy the condition 
$\Vert \textbf{x}_{i}^{t}-\textbf{y}_{m}^{t-1} \Vert < r_{vis}$, where $r_{vis}$ represents the visibility threshold. 
A state node $\textbf{y}_{m}^{t}$ is classified as visible if $q\geq V_{lim}$, where $V_{lim}$ is a predefined threshold; 
otherwise, it is marked as occluded. To obtain the visible state nodes position $\textbf{Y}_{vis}^{t}$, UPETrack aligns them  
to the current observation $\textbf{X}^{t}$ by registering $\textbf{Y}^{t-1}$ (as the initial GMM representation) 
to $\textbf{X}^{t}$ via EM algorithm.
In this context, each $\textbf{x}^{t}\in \textbf{X}^{t}$ is regarded as a sample generated from the Gaussian mixture distribution, 
while each $\textbf{y}^{t}\in \textbf{Y}^{t}$ corresponds to the mean of a distinct Gaussian component within the mixture. 
The mixture model assumes all Gaussian components share identical isotropic variances $\sigma ^{2}$ and a membership probability of $\frac{1}{M}$.
To enhance robustness against outliers and sensor noise, an additional uniform distribution component (indexed as $M+1$) with weight
$w$ is integrated into the mixture. The resulting joint probability density function of the GMM can then be expressed as follows:
\begin{align}
        p\left(\textbf{X}^t\right) =\prod_{n=1}^N{\sum_{m=1}^{M+1}{p\left( m \right) p\left( \textbf{x}_{n}^{t}\left| m \right. \right)}}
\end{align}
where
\begin{align}
p\left(\textbf{x}_{n}^{t}\left|m\right.\right) =\left\{ \begin{array}{l}
        \frac{1}{\left( 2\pi \sigma ^2 \right)^{\frac{D}{2}}}\exp \left( -\frac{\lVert \textbf{x}_{n}^{t}-\textbf{y}_m^t \rVert}{2\sigma ^2}\right) ,m=1,\cdots,M\\
        \frac{1}{N},\ \ \ \ \ \ \ \ \ \ \ \ \ \ \ \ \ \ \ \ \ \ \ \ \ \ \ \ \ m=M+1\\
\end{array} \right. 
\end{align}
\begin{align}
p\left( m \right) =\left\{ \begin{array}{l}
        \frac{1-\omega}{M},\ \ \ m=1,\cdots ,M\\
        \omega ,\ \ \ \ \ \ m=M+1\\
\end{array} \right. 
\end{align}

Within the E-step of GMM clustering, the posterior probabilities that maximize the data likelihood for 
the sampled points are computed through the application of Bayes' theorem, formulated as:
\begin{align}
p^{cur}\left( m\left| \textbf{x}_{n}^{t} \right. \right) =\left\{ \begin{array}{l}
        \frac{\exp \left( -\Delta _m \right)}{\sum\limits_{m=1}^M{\exp \left( -\Delta _m \right) +\mu}}, \ \ \ \ m=1,\cdots ,M\\
        \frac{1}{\frac{1}{\mu}\sum\limits_{m=1}^M\exp \left( -\Delta _m \right)+1}, \ \ m=M+1\\
\end{array} \right. 
\end{align}
where $\Delta _m= \frac{\lVert \textbf{x}_{n}^{t}-\textbf{y}_m^t \rVert^2}{2\sigma ^2}$, 
$\mu=\frac{\left( 2\pi \sigma ^2 \right) ^{\frac{D}{2}}\omega M}{\left( 1-\omega \right) N}$.

In the M-step, omitting items that are not related to the independent variable, the optimal $\textbf{Y}^t$ and 
$\sigma ^{2}$ can be obtained by maximizing the following cost function:
\begin{align}
Q \left( \textbf{Y}^t ,\sigma ^2 \right)=\sum_{n=1}^N{\sum_{m=1}^M {p^{cur} \left( m\left| \textbf{x}_n^t \right. \right)}}& \left( -\frac{D}{2} \log\sigma ^2- \right. \notag\\
& \left. \frac{\lVert \textbf{x}_n^t-\textbf{y}_m^t \rVert ^2}{2\sigma ^2}\right)
\end{align}
The E-step and M-step are iterated until convergence.

\begin{figure}[t]
  \centering
  \includegraphics[scale=0.17]{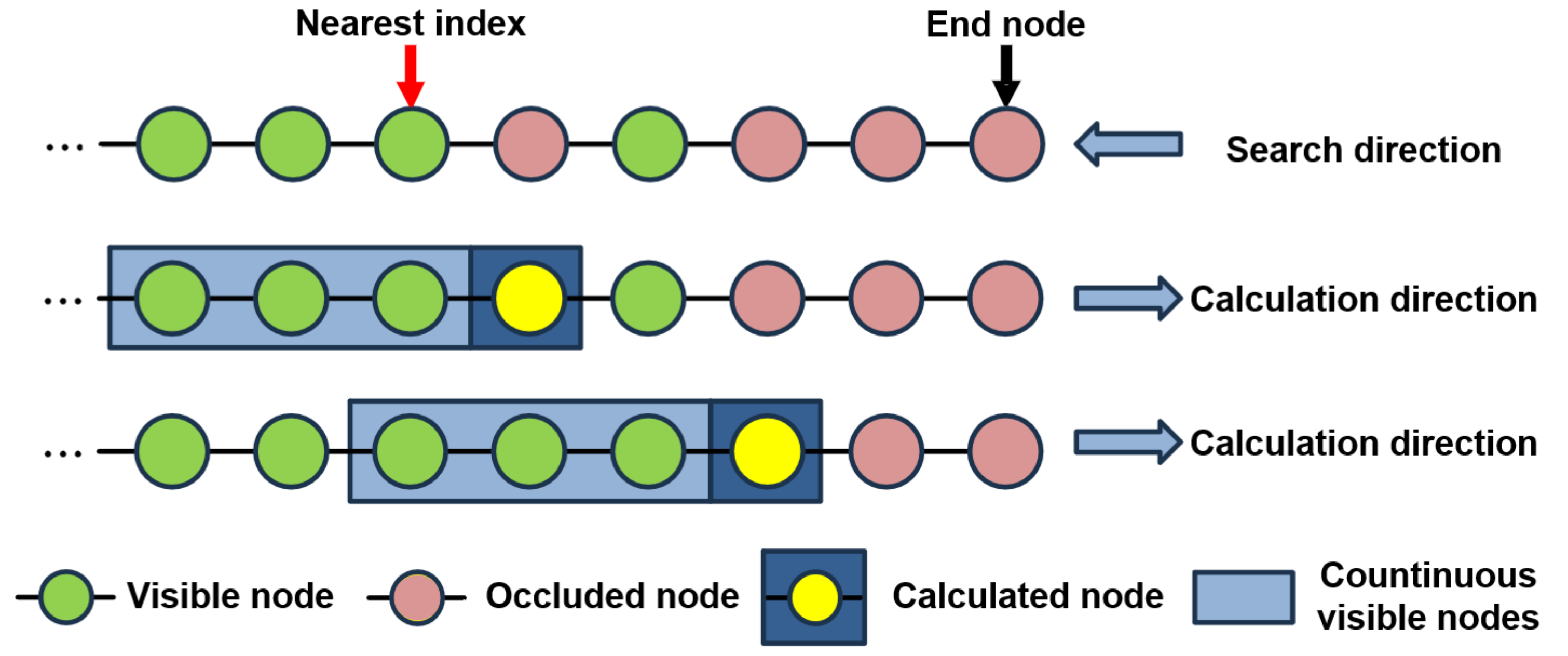}
  \caption{For tip occlusion scenarios, UPE traverses from the occluded end until locating a sequence of continuous 
  visible nodes. The algorithm then reverses direction, leveraging the continuous visible nodes to 
  iteratively estimate the nearest occluded node. Each estimated node is reclassified as visible, 
  propagating estimations until full occlusion resolution.}
  \label{figure}
  \end{figure}

The limitations of conventional GMM-EM-based tracking for partially occluded deformable linear objects (DLOs), 
as documented in prior studies \cite{ref23-10}, manifest as inaccuracies in shape estimation. To address this 
challenge, UPETrack employs a dual-strategy approach: visible nodes are tracked via the GMM-EM algorithm, while occluded nodes 
undergo estimation through the Unidirectional Position Estimation (UPE) algorithm, as detailed in Section III-B.

\subsection{Unidirectional Position Estimation} 
The UPE algorithm estimates the position of occluded nodes by leveraging the geometric continuity and temporal 
evolution patterns inherent in DLOs. Specifically, the positions of occluded nodes are derived from linear 
combinations of correlated node positions and displacements, based on the integration of three components: (1) a 
local linear combination displacement term, (2) a proximal linear 
constraint term, and (3) a historical curvature term.  
This approach significantly reduces computational complexity while maintaining tracking accuracy.

The UPE algorithm operates by utilizing the positional information of the continuous visible 
nodes closest to an occluded node to infer its position. The overall operational framework is 
illustrated in Fig. 4. The algorithm comprises two primary computational phases: (1) localization 
of nearest continuous visible nodes along the DLO, and (2) estimation of the occluded node position. 
In this work, the number of visible nodes in continuous visible nodes is set to 3.

In tip occlusion scenarios, the UPE algorithm initiates continuous visible nodes tracing from the occluded edge toward the opposite 
edge (this direction is defined as the \textbf{search direction}). Subsequently, utilizing the parameters of the continuous 
visible nodes, the coordinates of the occluded nodes are estimated in the direction opposite to the search direction (defined as the \textbf{calculation direction}, see Fig. 4).
Once the position of the occluded node is calculated, it is treated as a visible node and can be used to estimate the subsequent occluded node. 
This iterative process continues until the positions of all occluded nodes are determined. 
For mid-segment occlusion, the algorithm executes the search direction and calculation 
direction in both directions. The final positions of the occluded nodes are obtained by averaging 
the results from the two directions.

\begin{figure}[t]
  \centering
  \includegraphics[scale=0.16]{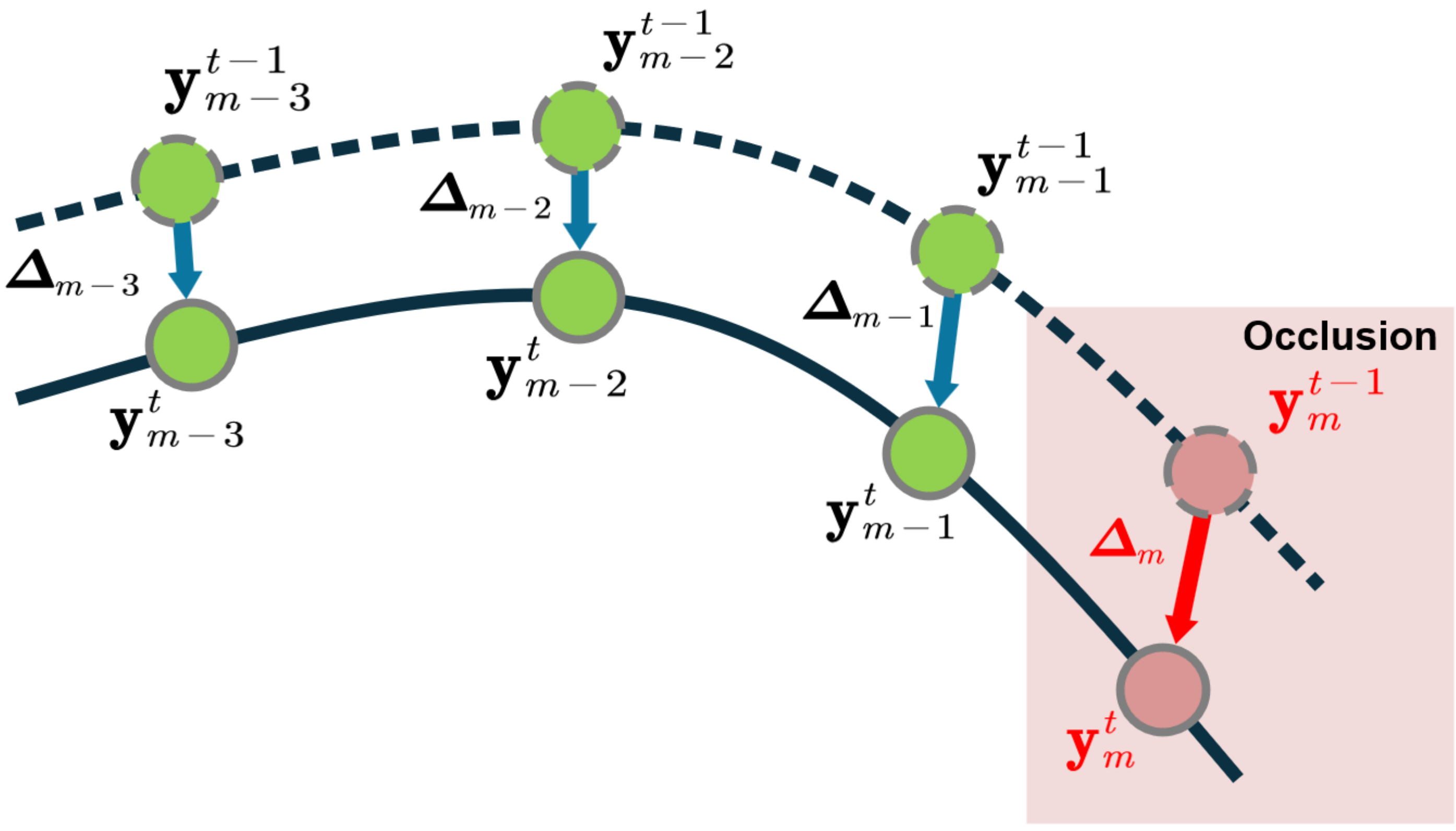}
  \caption{The local linear combination displacement represents the displacement $\boldsymbol{\varDelta} _{m}$ of the 
  occluded nodes $\textbf{y}_m^{t-1}$ as a distance-weighted sum of visible node displacements from the continuous visible nodes. }
  \label{figure}
  \end{figure}

A critical factor influencing both the accuracy and computational efficiency of UPE is the optimal configuration of the continuous visible nodes. 
To this end, the algorithm exploits the spatial continuity and temporal evolution characteristics of DLOs. 
Specifically, it integrates continuous visible nodes to generate a \textbf{local linear combination displacement term}, 
a \textbf{proximal linear constraint term}, and a \textbf{historical curvature term}; 
this framework enables efficient and accurate estimation of occluded node positions. 
Detailed mathematical formulations and explanations of these components are provided in the subsequent sections.

$Local$ $Linear$ $Combination$ $Displacement$ $Term$: The UPE algorithm assumes that the displacements of occluded nodes 
are strongly correlated with those of neighboring visible state nodes, with the strength of this correlation inversely proportional to the spatial distances between nodes. 
As depicted in Fig.5, let $\boldsymbol{\varDelta} _{i}$ denote the displacement vector of the visible state node $\textbf{y}_i$ 
across adjacent time steps, and let $\textbf{y}_m^{t}$ represent the occluded node to be estimated at time step $t$. 
This spatial correlation is modeled using a locally linear combination framework, which is mathematically formulated as follows:
\begin{align}
  \boldsymbol{\varDelta} _{m}=&\frac{1}{1+\gamma +\gamma ^2}\left( \boldsymbol{\varDelta} _{m-1}+\gamma \boldsymbol{\varDelta} _{m-2}+\gamma ^2\boldsymbol{\varDelta} _{m-3} \right)
  \end{align}
  Where $\gamma \in [0,1]$ is the distance attenuation factor, which ensures diminished influence from distant 
  visible nodes. Coefficient normalization is implemented through prefactor components in parentheses. 
  Then, the position vector of the occluded node can be obtained:
  \begin{align}
    \textbf{y}_m^{t'}=\textbf{y}_m^{t-1}+\boldsymbol{\varDelta} _m
          \end{align}

  \begin{figure}[t]
    \centering
    \includegraphics[scale=0.16]{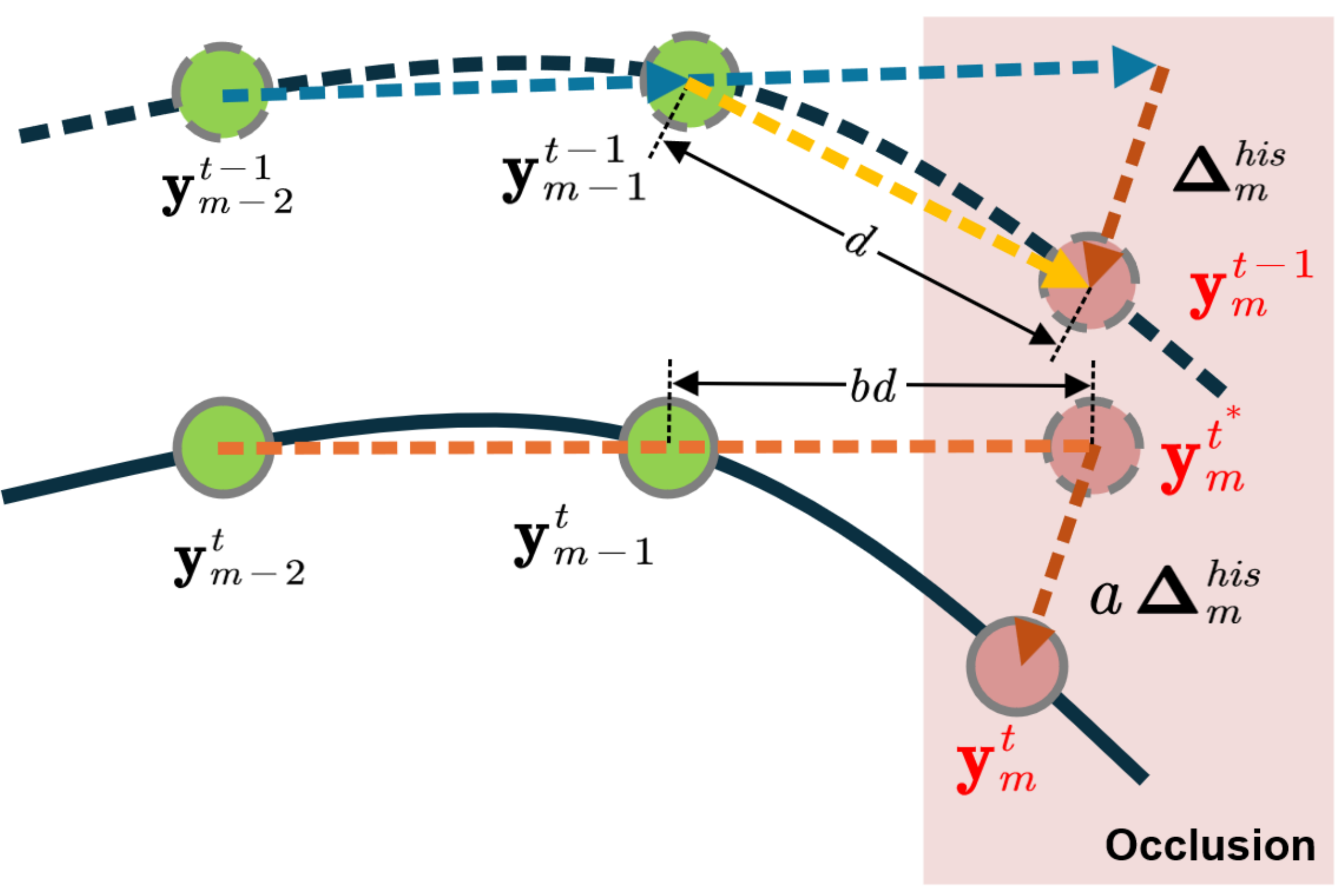}
    \caption{The proximal linear constraint term connects two visible nodes nearest to $\textbf{y}_{m}^{t}$,
    projecting a geometrically constrained line segment to define $\textbf{y}_{m}^{t*}$. The final position $\textbf{y}_{m}^{t*}$ 
    integrates this constraint with the historical bending term $\boldsymbol{\Delta }_{m}^{his}$ from prior states 
    through linear superposition.}
    \label{figure}
    \end{figure}
    
  $Proximal$ $Linear$ $Constraint$ $Term$: All DLOs exhibit inherent flexural rigidity, which helps maintain a straight-line 
  configuration between neighboring segments. As illustrated in Fig.6, where $d = \lVert \textbf{y}_{m}^{t-1}-\textbf{y}_{m-1}^{t-1}\rVert$, 
  this proximal geometric constraint can be analytically formulated as:
  \begin{align}
    \textbf{y}_{m}^{t^*}=\textbf{y}_{m-1}^{t}+b \lVert \textbf{y}_{m}^{t-1}-\textbf{y}_{m-1}^{t-1}\rVert \frac{\textbf{y}_{m-2}^{t}-\textbf{y}_{m-1}^{t}}{\lVert \textbf{y}_{m}^{t}-\textbf{y}_{m-1}^{t} \rVert} 
  \end{align}
  Here, the proximal rectilinear constraints term $\textbf{y}_{m}^{t^*}$ is computed based on the line formed 
  by the two nearest visible nodes in the continuous visible nodes. The scalar parameter $b$ is a bending resistance coefficient, 
  with larger values of $b$ indicating greater resistance to flexural deformation.

  $Historical$ $Curvature$ $Term$: When the time step is sufficiently small, DLOs exhibit minimal morphological changes between consecutive frames, 
  exhibiting strong temporal continuity in their geometric configurations. As shown in Fig. 6, to characterize this continuity, the UPE algorithm 
  constructs local tangent vectors from adjacent nodes and derives bending vectors by differencing the adjacent tangent vectors. The bending vectors 
  at the current time step are directly obtained from those at the previous time step. Mathematically, the historical curvature term 
  quantifying this temporal geometric continuity is defined as:
  \begin{align}
  \boldsymbol{\Delta }_{m}^{his}=\left(\textbf{y}_{m-1}^{t-1}-\textbf{y}_{m}^{t-1}\right)-\left(\textbf{y}_{m-2}^{t-1}-\textbf{y}_{m-1}^{t-1}\right) 
  \end{align}

  By integrating (8) and (9), the other estimated position of the occluded node can be derived:
  \begin{align}
    \textbf{y}_{m}^{t''}=\textbf{y}_{m}^{t^*}+a \boldsymbol{\Delta }_{m}^{his}
  \end{align}
  Coefficient $a$ governs the extent to which the bending configuration from the previous timestep is retained. 
  A higher value of $a$ preserves a greater proportion of the prior bending state, whereas a lower value retains less.

 \begin{figure}[t]
  \centering
  \begin{subfigure}[]{
    \includegraphics[scale=0.15]{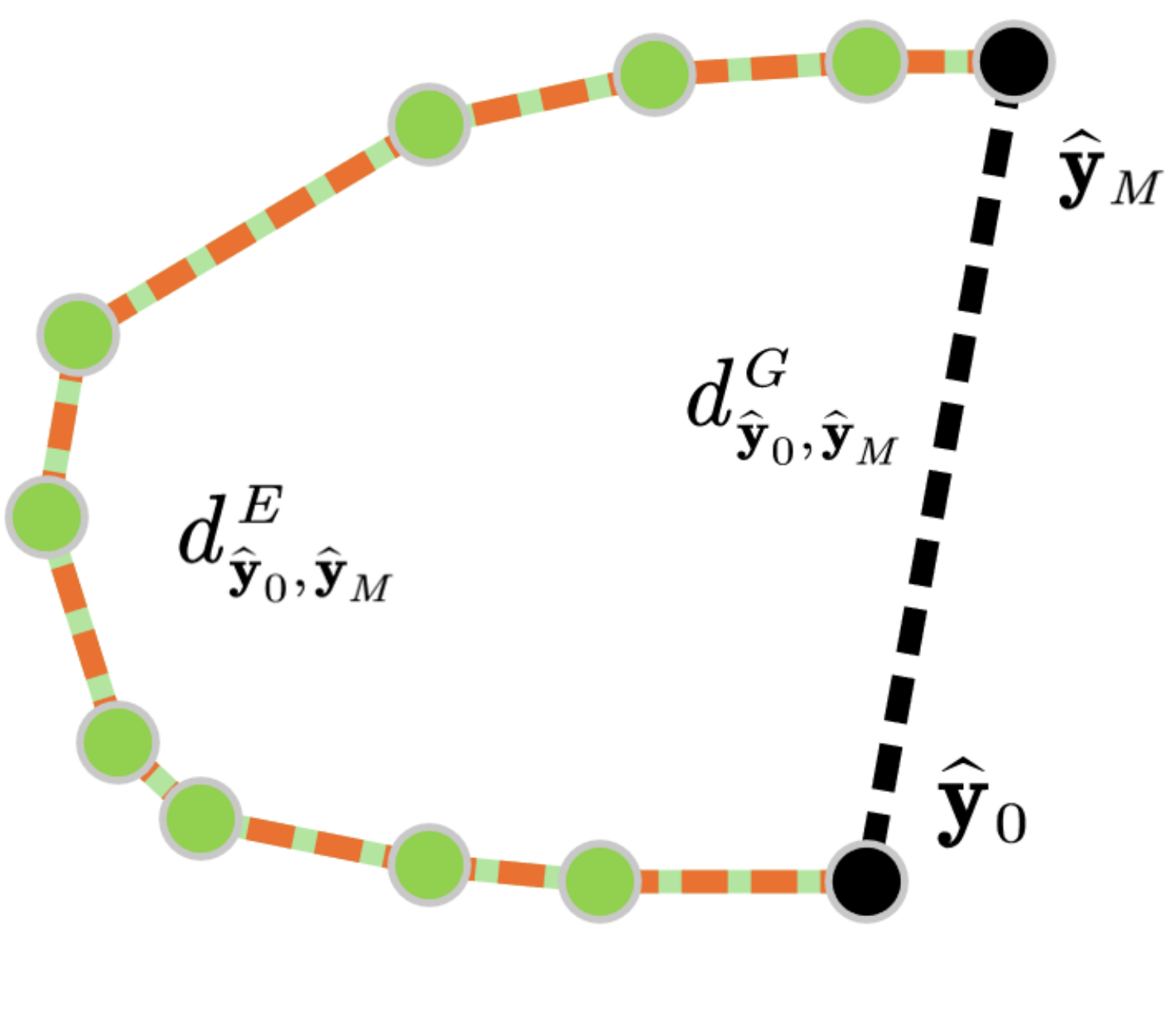}}
    \end{subfigure}
  \begin{subfigure}[]{
    \includegraphics[scale=0.22]{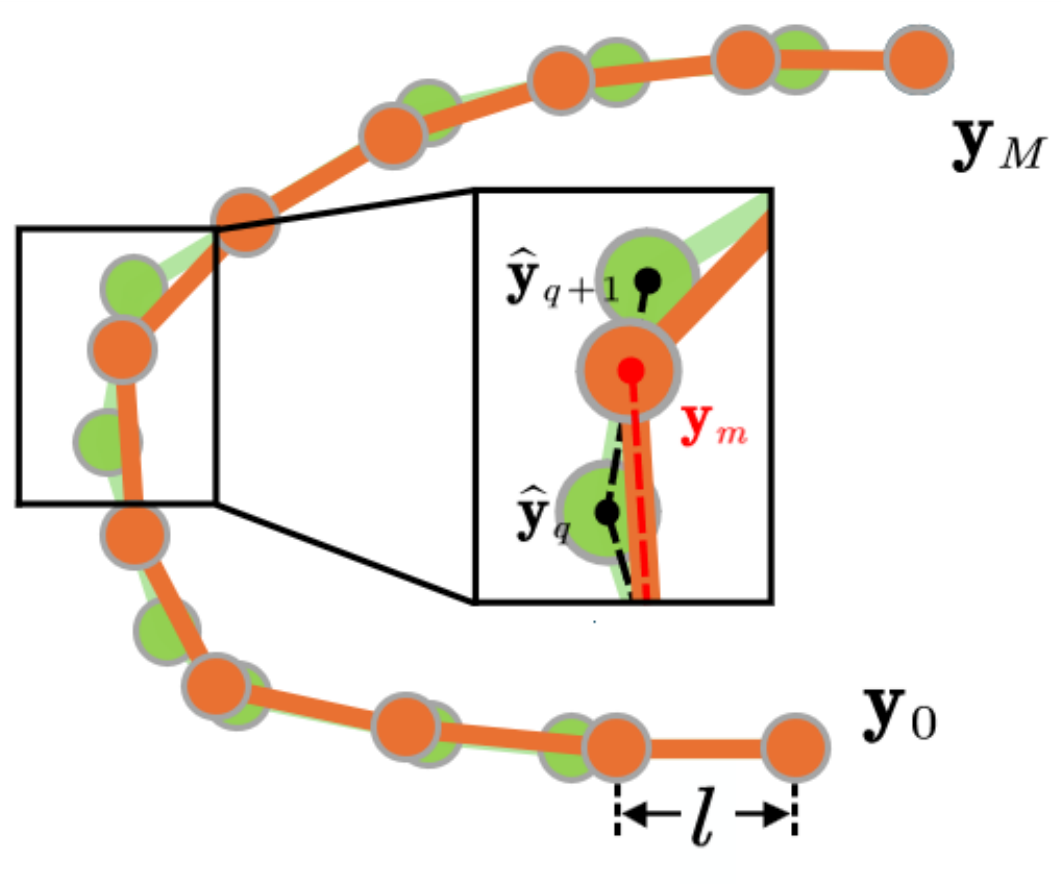}}
    \end{subfigure}
  \caption{(a) In the tracking algorithm for DLO, the geodesic distance (orange) 
    provides a more accurate representation of the distance 
    between two nodes compared to the Euclidean distance (black). 
    (b) Schematic diagram of the relationship between the resampling node $\textbf{y}_m$ and its related 
    parameter $q$ during the uniformization process.}
    \label{Fourier_Excitation_Advantage}
\end{figure}

  The final position estimation of occlusion nodes can be obtained by weighting the position estimations obtained by 
  equations (7) and (10):
  \begin{align}
    \textbf{y}_{m}^{t}=\alpha \textbf{y}_{m}^{t'}+(1-\alpha ) \textbf{y}_{m}^{t''} 
          \end{align}
  The weighting coefficient $\alpha$ governs the proportional contribution of two distinct computational methods.

  \begin{figure*}[t]
              \centering
              \includegraphics[width=1\textwidth]{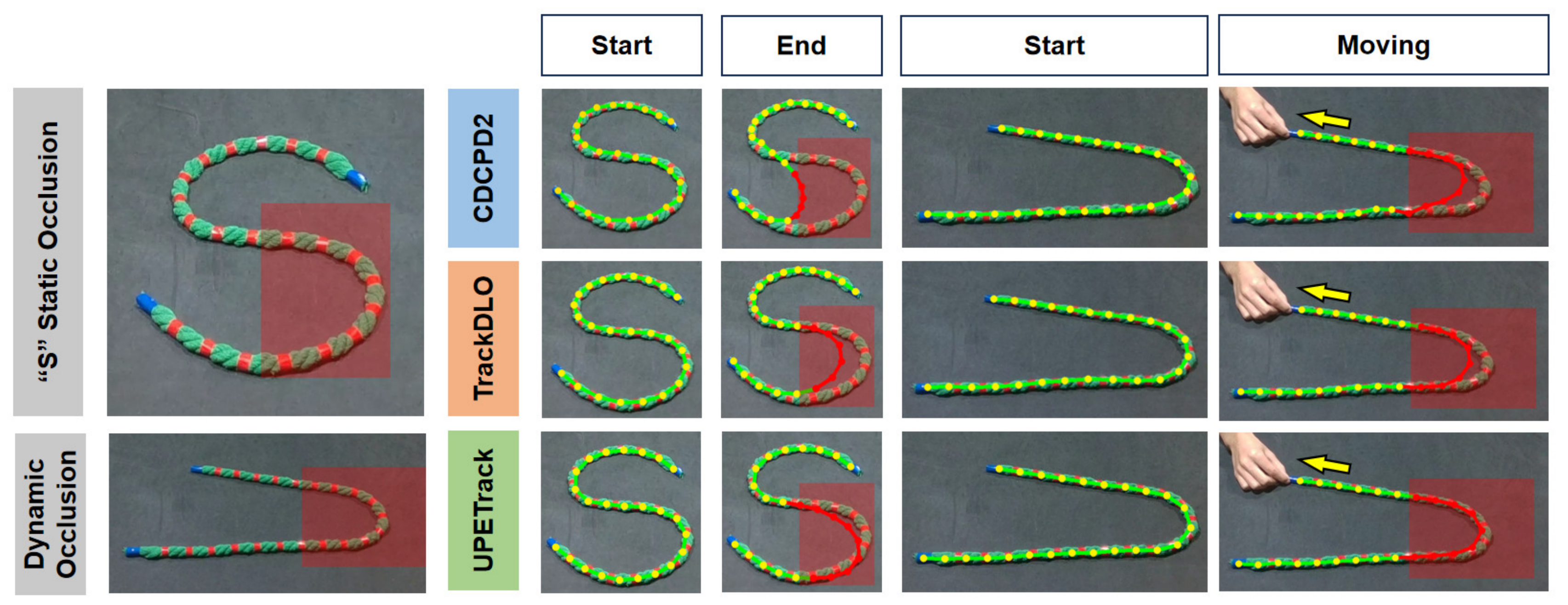}
              \caption{The comparative analysis of detection accuracy and efficiency in DLO tracking 
              is conducted for UPETrack, CDCPD2, and TrackDLO across two experimental scenarios: The 
              “S” static occlusion and the dynamic occlusion. The frame error 
              of the algorithm's detection of the DLO state position is calculated utilizing the red 
              and blue markers on the actual DLO as a reference for ground truth.}
              \end{figure*}

  The aforementioned estimation strategies are applicable when the continuous visible nodes are closer to the head 
  of the DLO (i.e., with indices smaller than that of the occluded node). When the continuous visible nodes are near 
  the end of the DLO (i.e., with indices greater than the occluded node), the UPE algorithm employs a similar strategy. 
  In this case, the continuous visible nodes are located on the other side of the occluded node. Consequently, the indices 
  $n-k$ (e.g., $n-1$, $n-2$) of the continuous visible nodes in the formulas are simply replaced with $n+k$ (e.g., $n+1$, $n+2$) 
  to accommodate this change.


\begin{figure}[t]
  \centering
  \begin{subfigure}[]{
    \includegraphics[scale=0.14]{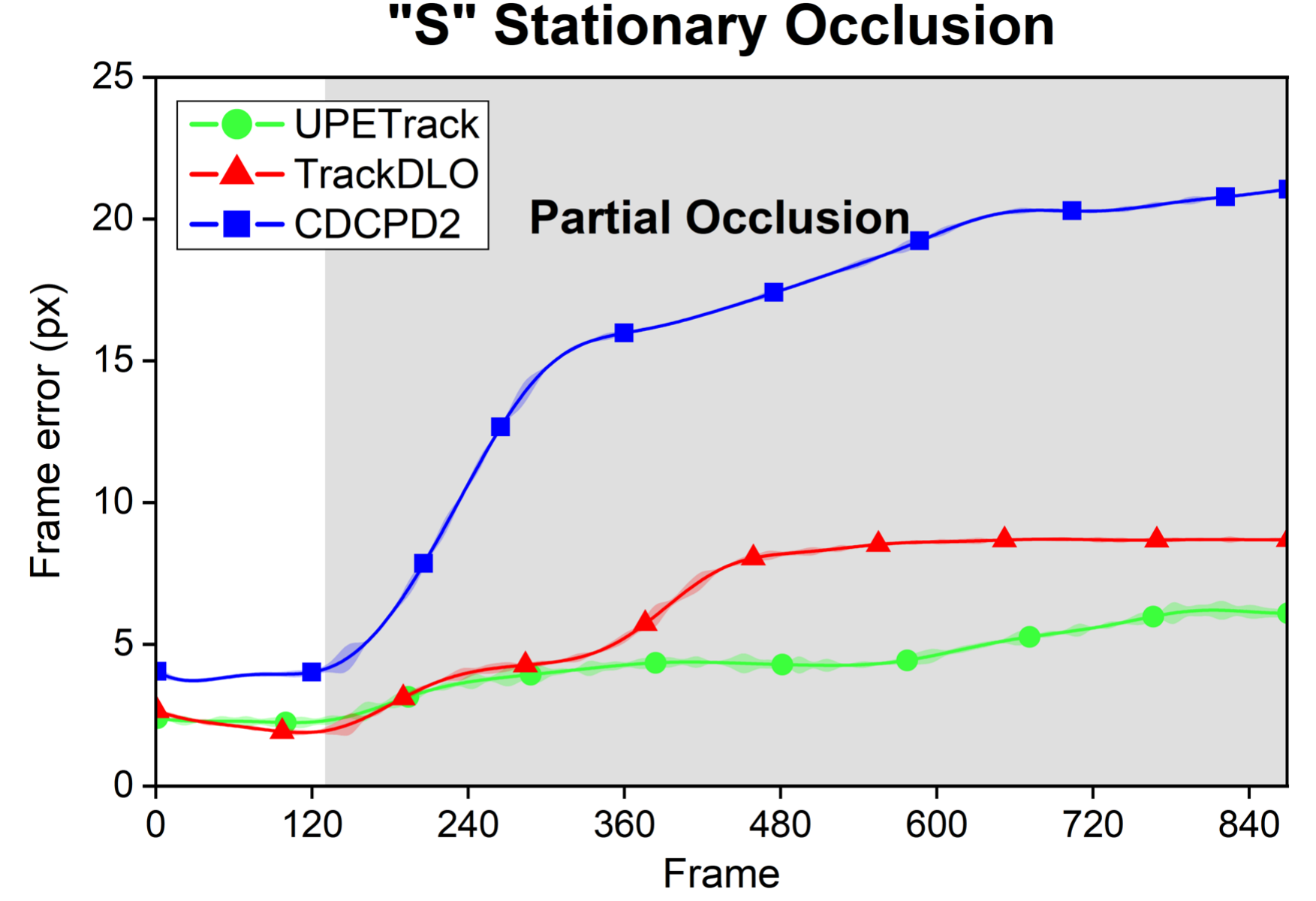}}
    \end{subfigure}
  \begin{subfigure}[]{
    \includegraphics[scale=0.14]{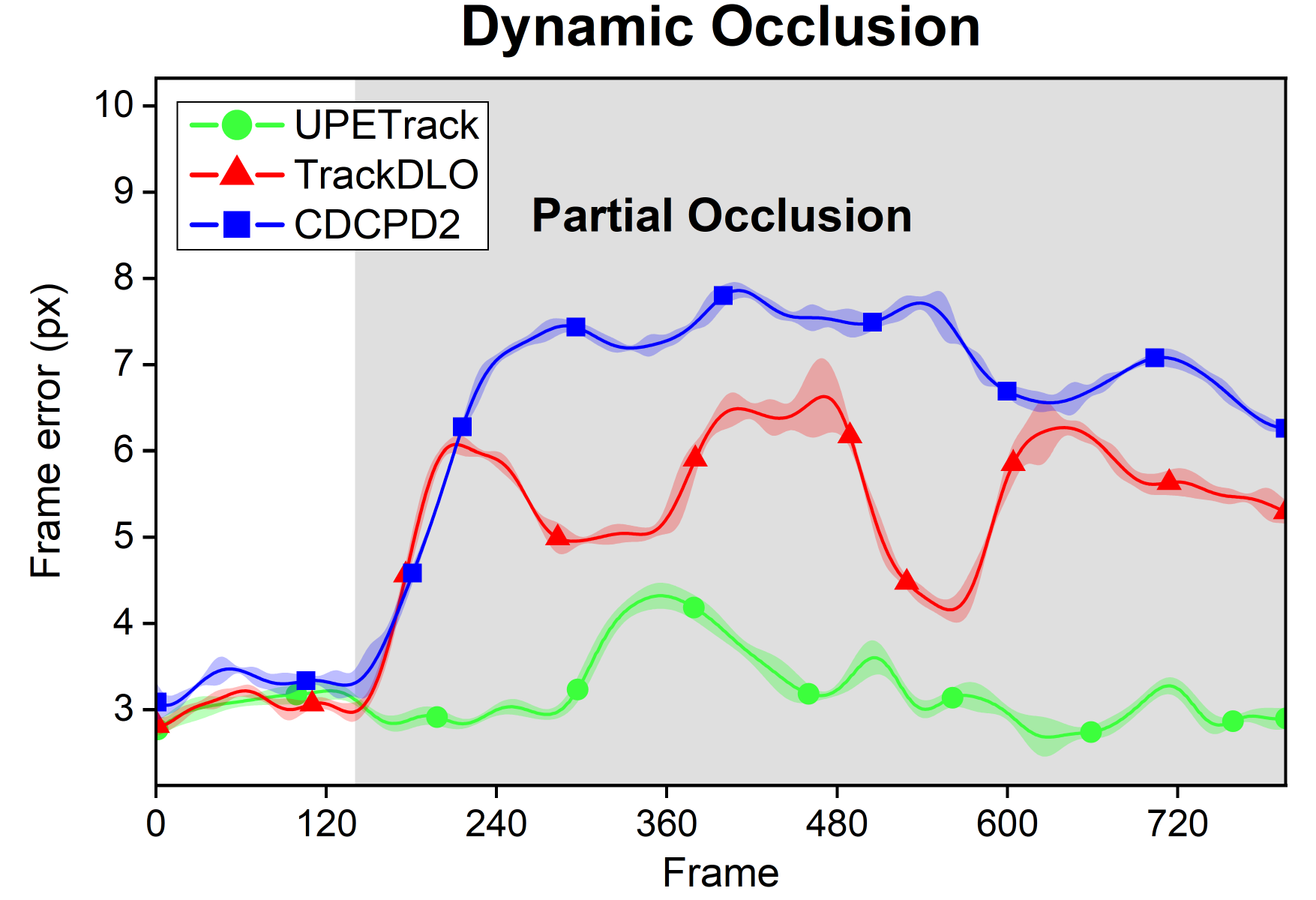}}
    \end{subfigure}
  \caption{(a)-(b) respectively illustrate the frame errors of three algorithms in 
                two experimental scenarios.}
  \label{Fourier_Excitation_Advantage}
\end{figure}

  \subsection{Nodes Distance Uniformization}

  The previously described methodologies for visible nodes tracking and occluded nodes localization 
  inherently struggle to maintain consistent inter-node distance. This limitation invariably leads to 
  uneven nodal distributions along the DLO axis. Such spatial non-uniformity progressively amplifies 
  numerical errors over successive tracking iterations, potentially culminating in tracking failure. 
  To mitigate this instability, a node-spacing uniformization resampling strategy is proposed and applied 
  to the original nodes obtained following each tracking iteration.  

  To achieve a more accurate characterization of inter-node spacing on deformable linear objects, 
  UPETrack abandons the Euclidean distance (depicted by black dashed lines in Fig. 7(a)), which 
  inadequately captures bending behavior in highly deformable structures, in favor of geodesic 
  distance (represented by orange dashed lines in Fig. 7(a)).

  The original nodes obtained after iteration are denoted as 
  $\widehat{\mathbf{Y}}=[\widehat{\textbf{y}}_{1},\widehat{\textbf{y}}_{2},\dots,\widehat{\textbf{y}}_{M}]$. 
  We order $\lVert \widehat{\textbf{y}}_0-\widehat{\textbf{y}}_1 \rVert=0$.
  Then, the geodesic distance constraints of the DLO between nodes $\widehat{\textbf{y}}_{j}^{t}$ and $\widehat{\textbf{y}}_{k}^{t}$ $(M \ge k > j \ge 1)$ can be calculated:
  \begin{align}
          d_{\widehat{\textbf{y}}_{j}^{t},\widehat{\textbf{y}}_{k}^{t}}^{G}=\sum_{i=j}^{k}{\lVert \widehat{\textbf{y}}_{i-1}^{t}-\widehat{\textbf{y}}_{i}^{t} \rVert} 
  \end{align}
  Thus, the length of the whole DLO $L$ can be expressed as:
  \begin{align}
  L=d_{\widehat{\textbf{y}}_{1}^{t},\widehat{\textbf{y}}_{M}^{t}}^{G}
  \end{align}

 \begin{figure}[t]
  \centering
  \begin{subfigure}[]{
    \includegraphics[scale=0.042]{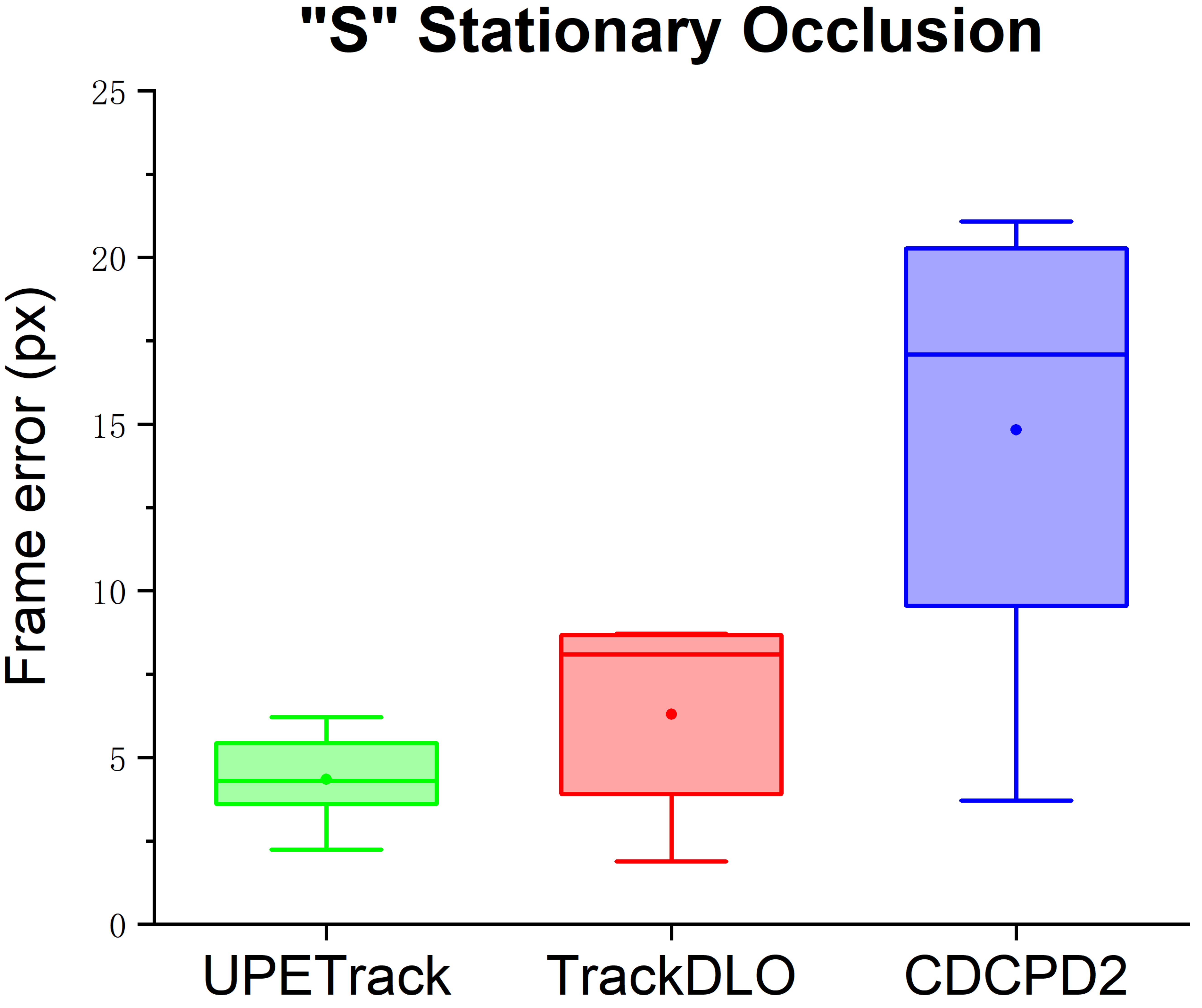}}
    \end{subfigure}
  \begin{subfigure}[]{
    \includegraphics[scale=0.042]{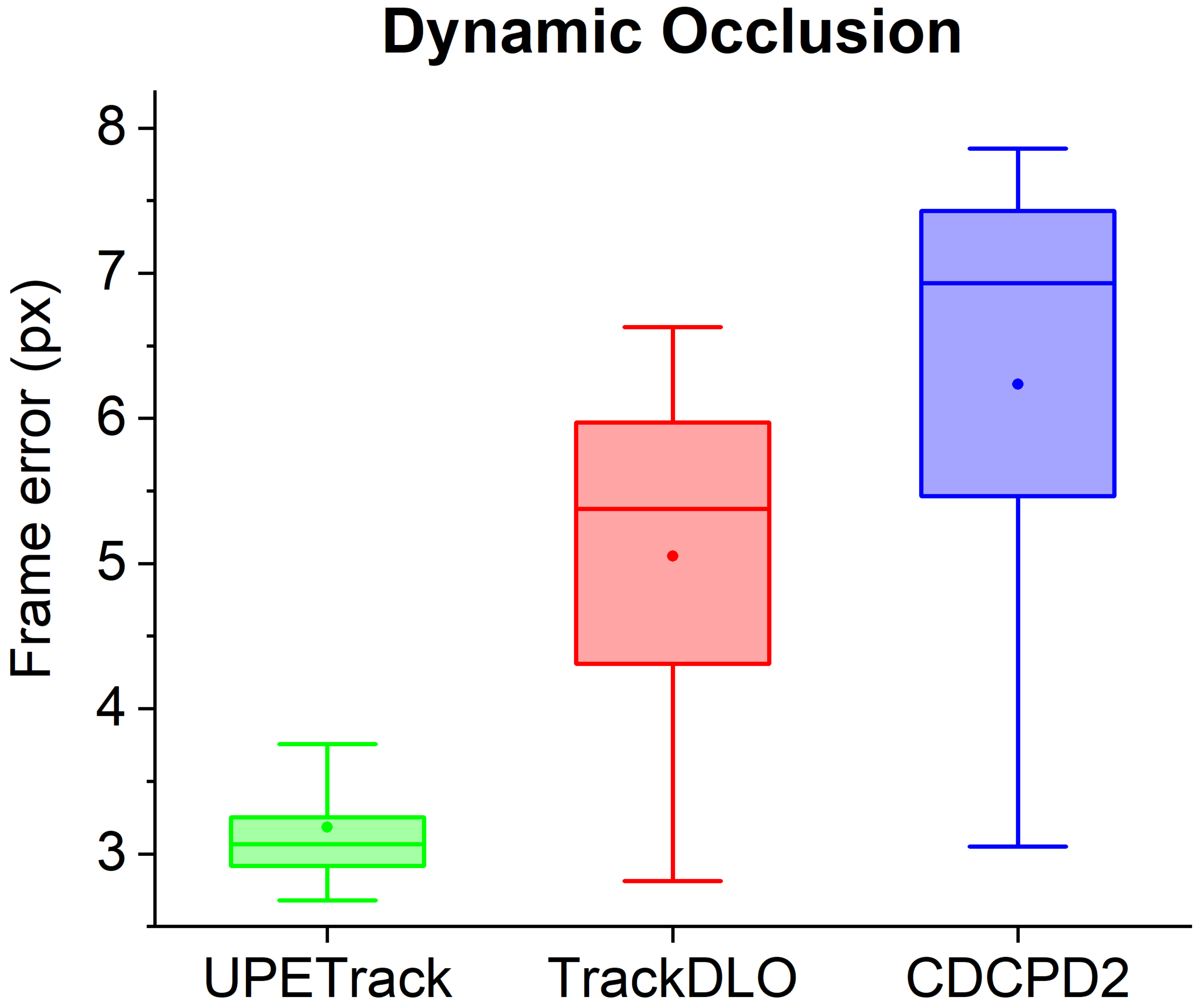}}
    \end{subfigure}
  \caption{(a)-(b) present comparative frame-error analyses under scenario I (“S”-shaped static occlusion) and II (Dynamic occlusion).}
  \label{Fourier_Excitation_Advantage}
\end{figure}

  The theoretical uniform length between the nodes can then be calculated as $l=\frac{L}{M-1}$.  As shown in the right panel of Fig.7(b), 
  the new state nodes $\textbf{Y}=[\textbf{y}_{1},\textbf{y}_{2},\dots,\textbf{y}_{M}]$ 
  will be obtained according to the following rules:

  \begin{align}
          \textbf{y}_{m}^{t}=\left\{\begin{array}{l}
          \widehat{\textbf{y}}_{1}^{t}\ \ \ \ \ \ \ \ \ \ \ \ \ \ \ \ \ \ \ \ \ \ \ \ \ \ \ \ \ \ \ \ \ \ \ \ \ \ \ m=1\\
          \widehat{\textbf{y}}_{q}^{t}-\frac{\left[ \left( n-1 \right) l-d_{\widehat{\textbf{y}}_{1}^{t},\widehat{\textbf{y}}_{q}^{t}}^{G} \right]}{\lVert \widehat{\textbf{y}}_{q}^{t}-\widehat{\textbf{y}}_{q+1}^{t} \rVert}\left(\widehat{\textbf{y}}_{q}^{t}-\widehat{\textbf{y}}_{q+1}^{t}\right)\ \ \ m\ne 1\\
  \end{array} \right. 
  \end{align}
  Where $q$ satisfies:  
  \begin{align}
  \left\{ \begin{array}{l}
          \left( n-1 \right) l-d_{\widehat{\textbf{y}}_{1}^{t},\widehat{\textbf{y}}_{q}^{t}}^{G} \ \ \ge 0\\
          \left( n-1 \right) l-d_{\widehat{\textbf{y}}_{1}^{t},\widehat{\textbf{y}}_{q+1}^{t}}^{G} < 0\\
  \end{array} \right. 
  \end{align}

  In this section, the UPETrack framework was evaluated against two state-of-the-art algorithms, TrackDLO and CDCPD2, focusing on tracking accuracy under occlusion  
  and computational efficiency for DLOs. The experimental data, including videos and point cloud, 
  were sourced from a RealSense D435i device. The UPETrack, TrackDLO, and CDCPD2 adopt fundamentally distinct 
  methodologies for occluded node estimation, resulting in distinct parameter configurations. 
  Specifically, the parameter configurations were set as follows: the proposed UPETrack  
  is configured with $M=24$, $\gamma=0.8$, $a=0.6$, $b=0.8$, $\alpha=0.75$, $r_{\text{vis}}=20$, and $V_{\text{lim}}=3$, whereas the 
  TrackDLO and CDCPD2 are configured with $M=24$, $k_{\text{vis}}=50$, $\beta=0.8$, $\lambda=120,\!000$, 
  $\alpha=3$, $\mu=0.5$, $d_{\text{vis}}=6$, and $\tau_{\text{vis}}=8$. The definitions of the parameters for the TrackDLO 
  and CDCPD2 algorithms strictly follow \cite{ref23-10} \cite{ref24-11}.

\section{EXPERIMENTS}

  \subsection{DLO-Tracking Accuracy}
  To evaluate tracking precision under occlusion, we conducted comparative 
  experiments on the three aforementioned frameworks using a cotton thread 
  (800 mm length $\times$ 15 mm diameter). 
  Two representative scenarios were designed to benchmark occlusion robustness:

  \begin{itemize}
          
          \item “S”-shaped Static Occlusion: 
          The DLO is initially arranged in an “S” shape, as shown in Fig. 8. The tracking 
          algorithm begins under unobstructed conditions. At a specific time point, an occlusion 
          is introduced at the inferior curvature of the “S” shape, while maintaining the overall 
          configuration of the DLO. This occlusion remains in place for the duration of the experiment.

          \item Dynamic Occlusion: 
          The DLO is configured as depicted in Fig. 8, with the tracking algorithm initialized 
          under unobstructed conditions. At a specific time point, one end of the DLO is pulled 
          to induce deformation, while complete occlusion is applied to the flexion region. This 
          occlusion is sustained for the remainder of the trial.

          \end{itemize}

  To evaluate tracking performance, we adopt the frame error metric as introduced in \cite{ref24-11}. 
  Given an ordered node sequence $\mathbf{Y}$ we define its corresponding piecewise linear (PWL) 
  curve as the set of all points on the straight-line segments connecting adjacent nodes in $\mathbf{Y}$, 
  including the nodes themselves. This set is denoted as PWL$(\mathbf{Y})$.

  The point-to-set distance between a point $\mathbf{y}$ and a set of points $\mathcal{M}$ is defined as:
  \begin{align}
          d\left( \mathbf{y,}\mathcal{M} \right) =\underset{\mathbf{y'}\in \mathcal{M}}{\text{inf}}\lVert \mathbf{y}-\mathbf{y'} \rVert 
  \end{align}
  At time $t$, the per-node error between the estimated state of the DLO $\mathbf{Y}^t$ and its ground-truth state $\mathbf{Y}_{\text{true}}^{t}$ is computed as:
  \begin{align}
          \varepsilon \left( \mathbf{Y}^t,\mathbf{Y}_{\text{true}}^{t} \right) =\frac{1}{M}\sum_{\mathbf{y}_{i}^{t}\in \mathbf{Y}^t}{d\left( \mathbf{y}_{i}^{t},\text{PWL}\left( \mathbf{Y}_{\text{true}}^{t} \right) \right)}
  \end{align} 
  To ensure symmetry in the evaluation, the final frame error metric is defined as the average of the bidirectional per-node errors:
  \begin{align}
          \mathcal{E}\left( \mathbf{Y}_{\text{true}}^{t},\mathbf{Y}^t \right) =\frac{1}{2}\left( \varepsilon \left( \mathbf{Y}_{\text{true}}^{t},\mathbf{Y}^t \right) +\varepsilon \left( \mathbf{Y}^t,\mathbf{Y}_{\text{true}}^{t} \right) \right) 
  \end{align}

  For each experimental scenario, multimodal sensor data, including video streams, point clouds, and associated metadata published by the RealSense node, were 
  stored in a Robot Operating System (ROS) bag file. Occlusions in the experiments were introduced by masking the raw input data within a specified region. 
  Specifically, for UPETrack and TrackDLO processing RGB-D data, occlusions were applied through color-channel masking in the corresponding regions of 
  the color images. For CDCPD2, which processes point cloud data, occlusions were implemented via 3D coordinate-based filtering of the point cloud. 
  The ground-truth positions for accuracy evaluation were acquired by detecting physical markers attached to the DLO. 
  Note these markers were solely used for ground truth acquisition and remained undetectable by all tracing algorithms. 
  Each experimental result was obtained from five repeated trials conducted under corresponding experimental scenarios.

  In the static "S-shaped occlusion" scenario, after initial 120-140 frames of unobstructed detection for DLO initialization, 
  continuous occlusion was applied to the lower-right bending region until the end of the experiment. 
  As shown in Fig. 9(a), the experimental results demonstrate that UPETrack achieves 
  the lowest steady-state error. Specifically, Fig. 10(a) reveals that throughout the experiment, UPETrack maintains 
  an interquartile range (IQR) of frame error between 3.6 and 5.4 px, significantly lower than both TrackDLO (3.9-8.7 px) 
  and CDCPD2 (9.6-20.2 px). The mean frame error of UPETrack is 4.3 px, representing reductions of 32\% and 70.8\% 
  compared to TrackDLO (6.3 px) and CDCPD2 (14.8 px), respectively.

  In the dynamic occlusion scenario, following the initial 120-140 frames of unobstructed detection, 
  complete occlusion was applied to the bending region prior to DLO motion. The DLO was then displaced by endpoint traction until the 
  experimental cycle concluded. Error analysis in Fig. 9(b) indicates that UPETrack achieved superior tracking 
  accuracy under dynamic occlusion. As demonstrated by Fig. 10(b), UPETrack achieved mean frame error of 3.18 px, 
  corresponding to reductions of 37.0\% and 48.9\% compared to TrackDLO (5.05 px) and CDCPD2 (6.23 px), respectively. 
  Its IQR (2.9-3.3 px) was consistently lower than that of both TrackDLO (4.3-6.0 px) and CDCPD2 (5.5-7.4 px). 
  These results confirm UPETrack is capable of accurately reconstructing the states of occluded DLO segments, 
  demonstrating enhanced robustness to occlusion when compared to TrackDLO and CDCPD2.

  \begin{table}[t]
        \caption{Tracking Time Comparison}
        \label{table_example}
        \begin{center}
        \begin{tabular}{|c|c|c|c|}
        \hline
        Algorithm & CDCPD2 & TrackDLO & \textbf{UPETrack(Ours)}\\
        \hline
        Tracking Time (ms) & 16.495 & 14.731 & \textbf{11.72}\\
        \hline
        \end{tabular}
        \end{center}
        \end{table}

\subsection{DLO-Tracking Efficiency}

  To assess computational efficiency, we compared the per-frame processing times of UPETrack, 
  TrackDLO, and CDCPD2 under both static "S-shaped occlusion" and dynamic occlusion scenarios. 
  The experiments were conducted using ROS dataset files, with each algorithm executed independently 
  for 10 trials to calculate the mean processing time and standard deviation. The tests were run on a 
  hardware platform consisting of an Ubuntu 20.04 system with an Intel Core i7-9700 CPU and an NVIDIA 
  GeForce RTX 4060 Ti GPU. TrackDLO and CDCPD2 were implemented in C++, whereas UPETrack was developed in Python. 
  Timing was strictly limited to the core tracking algorithm, spanning from valid input reception to output 
  generation. Non-core processes (ROS communication, segmentation, point cloud generation, downsampling, 
  occlusion handling) were excluded. As shown in Table 1, UPETrack achieved a mean processing time per frame of 
  79.6\% and 71.1\% of the times observed for TrackDLO and CDCPD2, respectively. 

\section{Conclusion and Future Work}
  This paper proposes UPETrack, a geometry-driven framework for real-time visual tracking 
  of DLOs under partial occlusion. Experimental evaluations show that UPETrack outperforms 
  TrackDLO and CDCPD2 in tracking accuracy and efficiency. 
  However, UPETrack exhibits limitations including 
  adverse impact of endpoint occlusion on tracking accuracy, 
  susceptibility to tracking failures under large inter-frame displacements. 
  Future research will explore the deployment of UPETrack in real-world robotic applications, particularly in tasks requiring precise DLO shape control and manipulation \cite{robotics13010018}, \cite{10993307}.
  

%


\ifCLASSOPTIONcaptionsoff
  \newpage
\fi



%

\bibliographystyle{ieeetran}

\bibliography{mybibfile}

@inproceedings{ref10,
  author    = {Shah, Ankit J. and Shah, Julie A.},
  booktitle = {2016 IEEE International Conference on Robotics and Automation (ICRA)},
  title     = {Towards manipulation planning for multiple interlinked deformable linear objects},
  year      = {2016},
  volume    = {},
  number    = {},
  pages     = {3908-3915},
  doi       = {10.1109/ICRA.2016.7487580}
}

@article{refWang2023,
  author    = {Wang, Xi and Zhu, Yifan},
  journal   = {Machines}, 
  title     = {Deformable Object Manipulation in Caregiving Scenarios: A Review},
  year      = {2023},
  volume    = {},
  number    = {},
  pages     = {1013},
  doi       = {10.3390/machines11111013},
  publisher = {MDPI}
}

@inproceedings{10341726,
  author={Chen, Kejia and Bing, Zhenshan and Wu, Fan and Meng, Yuan and Kraft, André and Haddadin, Sami and Knoll, Alois},
  booktitle={2023 IEEE/RSJ International Conference on Intelligent Robots and Systems (IROS)}, 
  title={Contact-Aware Shaping and Maintenance of Deformable Linear Objects With Fixtures}, 
  year={2023},
  volume={},
  number={},
  pages={1-8},
  doi={10.1109/IROS55552.2023.10341726}}

@inproceedings{ref11,
  author    = {Ge, Song and Fan, Guoliang and Ding, Meng},
  booktitle = {2014 IEEE Conference on Computer Vision and Pattern Recognition Workshops},
  title     = {Non-rigid Point Set Registration with Global-Local Topology Preservation},
  year      = {2014},
  volume    = {},
  number    = {},
  pages     = {245-251},
  doi       = {10.1109/CVPRW.2014.45}
}

@article{ref12,
  author    = {H. Yin and A. Varava and D. Kragic},
  journal   = {Science Robotics},
  title     = {Modeling, learning, perception, and control methods for deformable object manipulation},
  year      = {2021},
  volume    = {6},
  number    = {},
  pages     = {1-16},
  month     = {May},
  doi       = {},
  issn      = {},
  publisher = {}
}

@article{ref13,
  author   = {Dempster, A. P. and Laird, N. M. and Rubin, D. B.},
  title    = {Maximum Likelihood from Incomplete Data Via the EM Algorithm},
  journal  = {Journal of the Royal Statistical Society: Series B (Methodological)},
  volume   = {39},
  number   = {1},
  pages    = {1-22},
  year     = {2018},
  month    = {12},
  issn     = {0035-9246},
  doi      = {10.1111/j.2517-6161.1977.tb01600.x},
  url      = {https://doi.org/10.1111/j.2517-6161.1977.tb01600.x},
  eprint   = {https://academic.oup.com/jrsssb/article-pdf/39/1/1/49117094/jrsssb\_39\_1\_1.pdf}
}

@INPROCEEDINGS{ref2,
  author={Galassi, Kevin and Palli, Gianluca},
  booktitle={2021 4th IEEE International Conference on Industrial Cyber-Physical Systems (ICPS)}, 
  title={Robotic Wires Manipulation for Switchgear Cabling and Wiring Harness Manufacturing}, 
  year={2021},
  volume={},
  number={},
  pages={531-536},
  doi={10.1109/ICPS49255.2021.9468128}}

@inproceedings{ref3,
  author    = {Lu, Bo and Chu, Henry K. and Cheng, Li},
  booktitle = {2016 IEEE International Conference on Real-time Computing and Robotics (RCAR)},
  title     = {Dynamic trajectory planning for robotic knot tying},
  year      = {2016},
  volume    = {},
  number    = {},
  pages     = {180-185},
  doi       = {10.1109/RCAR.2016.7784022}
}

@inproceedings{ref4,
  author    = {Keipour, Azarakhsh and Bandari, Maryam and Schaal, Stefan},
  booktitle = {2022 IEEE/RSJ International Conference on Intelligent Robots and Systems (IROS)},
  title     = {Efficient Spatial Representation and Routing of Deformable One-Dimensional Objects for Manipulation},
  year      = {2022},
  volume    = {},
  number    = {},
  pages     = {211-216},
  doi       = {10.1109/IROS47612.2022.9981939}
}

@article{ref5,
  title    = {An algorithm based on bidirectional searching and geometric constrained sampling for automatic manipulation planning in aircraft cable assembly},
  journal  = {Journal of Manufacturing Systems},
  volume   = {57},
  pages    = {158-168},
  year     = {2020},
  issn     = {0278-6125},
  doi      = {https://doi.org/10.1016/j.jmsy.2020.08.015},
  url      = {https://www.sciencedirect.com/science/article/pii/S0278612520301552},
  author   = {Jiuming Guo and Jiwen Zhang and Dan Wu and Yuhang Gai and Ken Chen}
}

@article{ref6,
  title       = {{Robotic Manipulation and Sensing of Deformable Objects in Domestic and Industrial Applications: A Survey}},
  author      = {Sanchez, Jose and Corrales Ramon, Juan Antonio and Bouzgarrou, Belhassen-Chedli and Mezouar, Youcef},
  url         = {https://uca.hal.science/hal-01816189},
  journal     = {{The International Journal of Robotics Research}},
  publisher   = {{SAGE Publications}},
  volume      = {37},
  number      = {7},
  pages       = {688 - 716},
  year        = {2018},
  month       = Jun,
  doi         = {10.1177/0278364918779698},
  hal_id      = {hal-01816189},
  hal_version = {v1}
}

@article{ref7,
  author={Haiderbhai, Mustafa and Gondokaryono, Radian and Wu, Andrew and Kahrs, Lueder A.},
  journal={IEEE Transactions on Automation Science and Engineering}, 
  title={Sim2Real Rope Cutting With a Surgical Robot Using Vision-Based Reinforcement Learning}, 
  year={2025},
  volume={22},
  number={},
  pages={4354-4365},
  doi={10.1109/TASE.2024.3410297}}

@article{ref8,
  title    = {Overview of the State of the Art in the Production Process of Automotive Wire Harnesses, Current Research and Future Trends},
  journal  = {Procedia CIRP},
  volume   = {81},
  pages    = {387-392},
  year     = {2019},
  note     = {52nd CIRP Conference on Manufacturing Systems (CMS), Ljubljana, Slovenia, June 12-14, 2019},
  issn     = {2212-8271},
  doi      = {https://doi.org/10.1016/j.procir.2019.03.067},
  url      = {https://www.sciencedirect.com/science/article/pii/S2212827119303725},
  author   = {Jerome Trommnau and Jens Kühnle and Jörg Siegert and Robert Inderka and Thomas Bauernhansl}
}

@ARTICLE{ref14-1,
  author={Myronenko, Andriy and Song, Xubo},
  journal={IEEE Transactions on Pattern Analysis and Machine Intelligence}, 
  title={Point Set Registration: Coherent Point Drift}, 
  year={2010},
  volume={32},
  number={12},
  pages={2262-2275},
  doi={10.1109/TPAMI.2010.46}}

@ARTICLE{ref15-2,
  author={A.L. Yuille and N.M. Grzywacz},
  journal={International Journal of Computer Vision},
  title={A mathematical analysis of the motion coherence theory},
  year={1989},
  volume={3},
  number={},
  pages={155-175},
  doi={10.1007/BF00126430},
  ISSN={},
  publisher={Springer}
}

@INPROCEEDINGS{ref16-3,
  author={Tang, Te and Fan, Yongxiang and Lin, Hsien-Chung and Tomizuka, Masayoshi},
  booktitle={2017 IEEE/RSJ International Conference on Intelligent Robots and Systems (IROS)}, 
  title={State estimation for deformable objects by point registration and dynamic simulation}, 
  year={2017},
  volume={},
  number={},
  pages={2427-2433},
  doi={10.1109/IROS.2017.8206058}}

@ARTICLE{ref17-4,
  author={Tang, Te and Wang, Changhao and Tomizuka, Masayoshi},
  journal={IEEE Robotics and Automation Letters}, 
  title={A Framework for Manipulating Deformable Linear Objects by Coherent Point Drift}, 
  year={2018},
  volume={3},
  number={4},
  pages={3426-3433},
  doi={10.1109/LRA.2018.2852770}}

@ARTICLE{ref18-5,
  author={Tang, T. and Tomizuka, M.},
  journal={The International Journal of Robotics Research},
  title={Track deformable objects from point clouds with structure preserved registration},
  year={2019},
  volume={41},
  number={6},
  pages={599-614},
  doi={10.1177/0278364919841431},
  ISSN={},
  publisher={}
}

@ARTICLE{ref20-7,
  author={Jin, Shiyu and Lian, Wenzhao and Wang, Changhao and Tomizuka, Masayoshi and Schaal, Stefan},
  journal={IEEE Robotics and Automation Letters}, 
  title={Robotic Cable Routing with Spatial Representation}, 
  year={2022},
  volume={7},
  number={2},
  pages={5687-5694},
  doi={10.1109/LRA.2022.3158377}}

@inproceedings{ref21-8,
author = {Chi, Cheng and Berenson, Dmitry},
title = {Occlusion-robust Deformable Object Tracking without Physics Simulation},
year = {2019},
publisher = {IEEE Press},
url = {https://doi.org/10.1109/IROS40897.2019.8967827},
doi = {10.1109/IROS40897.2019.8967827},
booktitle = {2019 IEEE/RSJ International Conference on Intelligent Robots and Systems (IROS)},
pages = {6443–6450},
numpages = {8},
location = {Macau, China}
}

@ARTICLE{ref22-9,
  author={Sam T. Roweis and Lawrence K. Saul},
  journal={Science},
  title={Nonlinear Dimensionality Reduction by Locally Linear Embedding},
  year={2000},
  volume={290},
  number={5500},
  pages={2323-2326},
  doi={10.1126/science.290.5500.2323},
  ISSN={},
  publisher={}
}

@INPROCEEDINGS{ref23-10,
  author={Y. Wang and D. McConachie and D. Berenson},
  booktitle={2021 IEEE International Conference on Robotics and Automation (ICRA)}, 
  title={Tracking Partially-Occluded Deformable Objects while Enforcing Geometric Constraints}, 
  year={2021},
  pages={14199-14205},
  doi={10.1109/ICRA48506.2021.9561012},
  ISSN={},
  publisher={IEEE}
}

@ARTICLE{ref24-11,
  author={Xiang, Jingyi and Dinkel, Holly and Zhao, Harry and Gao, Naixiang and Coltin, Brian and Smith, Trey and Bretl, Timothy},
  journal={IEEE Robotics and Automation Letters}, 
  title={TrackDLO: Tracking Deformable Linear Objects Under Occlusion With Motion Coherence}, 
  year={2023},
  volume={8},
  number={10},
  pages={6179-6186},
  doi={10.1109/LRA.2023.3303710}}

@INPROCEEDINGS{ref25-12,
  author={Zhang, Jiaming and Zhang, Zhaomeng and Liu, Yihao and Chen, Yaqian and Kheradmand, Amir and Armand, Mehran},
  booktitle={2024 IEEE International Conference on Robotics and Automation (ICRA)}, 
  title={Realtime Robust Shape Estimation of Deformable Linear Object}, 
  year={2024},
  volume={},
  number={},
  pages={10734-10740},
  doi={10.1109/ICRA57147.2024.10610432}}

@ARTICLE{ref26-13,
  author={Yang, Yuxuan and Stork, Johannes A. and Stoyanov, Todor},
  journal={IEEE Robotics and Automation Letters}, 
  title={Particle Filters in Latent Space for Robust Deformable Linear Object Tracking}, 
  year={2022},
  volume={7},
  number={4},
  pages={12577-12584},
  doi={10.1109/LRA.2022.3216985}}

@INPROCEEDINGS{ref27-14,
  author={K. Lv and M. Yu and Y. Pu and X. Jiang and G. Huang and X. Li},
  booktitle={Proc. IEEE Int. Conf. Robot. Automat. Workshop Representing Manipulating Deformable Objects}, 
  title={Learning to estimate 3-D states of deformable linear objects from single-frame occluded point clouds}, 
  year={2023},
  pages={7119-7125},
  doi={},
  ISSN={},
  publisher={IEEE}
}

@ARTICLE{ref28,
  author={Zhu, Jihong and Cherubini, Andrea and Dune, Claire and Navarro-Alarcon, David and Alambeigi, Farshid and Berenson, Dmitry and Ficuciello, Fanny and Harada, Kensuke and Kober, Jens and Li, Xiang and Pan, Jia and Yuan, Wenzhen and Gienger, Michael},
  journal={IEEE Robotics \& Automation Magazine}, 
  title={Challenges and Outlook in Robotic Manipulation of Deformable Objects}, 
  year={2022},
  volume={29},
  number={3},
  pages={67-77},
  doi={10.1109/MRA.2022.3147415}}

@ARTICLE{ref29,
  author={Caporali, Alessio and Galassi, Kevin and Zanella, Riccardo and Palli, Gianluca},
  journal={IEEE Robotics and Automation Letters}, 
  title={FASTDLO: Fast Deformable Linear Objects Instance Segmentation}, 
  year={2022},
  volume={7},
  number={4},
  pages={9075-9082},
  doi={10.1109/LRA.2022.3189791}}

@Article{robotics13010018,
AUTHOR = {Almaghout, Karam and Klimchik, Alexandr},
TITLE = {Manipulation Planning for Cable Shape Control},
JOURNAL = {Robotics},
VOLUME = {13},
YEAR = {2024},
NUMBER = {1},
ARTICLE-NUMBER = {18},
URL = {https://www.mdpi.com/2218-6581/13/1/18},
ISSN = {2218-6581},
DOI = {10.3390/robotics13010018}
}

@ARTICLE{10993307,
  author={Caporali, Alessio and Palli, Gianluca},
  journal={IEEE/ASME Transactions on Mechatronics}, 
  title={Robotic Manipulation of Deformable Linear Objects via Multiview Model-Based Visual Tracking}, 
  year={2025},
  volume={},
  number={},
  pages={1-12},
  doi={10.1109/TMECH.2025.3562295}}

%







\end{document}